\documentclass[journal]{IEEEtran}
\ifCLASSINFOpdf
\else
\fi
\usepackage{graphicx}

\usepackage{threeparttable}
\usepackage{subfigure}

\usepackage{multicol}
\usepackage{multirow}

\usepackage{CJK}

\usepackage{amsmath,amssymb,amsfonts}
\usepackage{booktabs}

\newcounter{rownumbers}

\usepackage{verbatim}
\usepackage{color}
\usepackage{cite}
\usepackage{ulem}
\usepackage[ruled, vlined, linesnumbered]{algorithm2e}

\usepackage{url}
\usepackage[colorlinks,linkcolor=red]{hyperref}
\begin{document}
\title{Temporal Knowledge Graph Reasoning Triggered by Memories}

\author{Mengnan~Zhao,
	Lihe~Zhang,
	Yuqiu~Kong$\dagger$,
	Baocai~Yin
}

\markboth{IEEE Transactions on Multimedia,~Vol.~xx, No.~x, Month~20xx}%
{Zhao \MakeLowercase{\textit{et al.}}: Temporal Knowledge Graph Reasoning Triggered by Memories.}

\maketitle

\begin{abstract}
Inferring missing facts in temporal knowledge graphs is a critical task and has been widely explored. Extrapolation in temporal reasoning tasks is more challenging and gradually attracts the attention of researchers since no direct history facts for prediction. 
Previous works attempted to apply evolutionary representation learning to solve the extrapolation problem. However, these techniques do not explicitly leverage various time-aware attribute representations and are not aware of the event dissolution process.
To alleviate the time dependence when reasoning future missing facts, we propose a Memory-Triggered Decision-Making (MTDM) network, which incorporates transient memories, long-short-term memories, and deep memories. Specifically, the transient learning network considers transient memories as a static knowledge graph, and the time-aware recurrent evolution network learns representations through a sequence of recurrent evolution units from long-short-term memories. Each evolution unit consists of a structural encoder to aggregate edge information, a time encoder with a gating unit to update attribute representations of entities. MTDM utilizes the crafted residual multi-relational aggregator as the structural encoder to solve the multi-hop coverage problem. 
For better understanding the event dissolution process, we introduce the dissolution learning constraint.
Extensive experiments demonstrate the MTDM alleviates the time dependence and achieves state-of-the-art prediction performance. Moreover, compared with the most advanced baseline, MTDM shows a faster convergence speed and training speed.

\end{abstract}

\section{Introduction}\label{sec:introduction}
\IEEEPARstart{R}{easoning} on Temporal Knowledge Graphs (TKGs) aggregate the time-aware real-world scenarios to infer missing facts and is applied in various tasks, such as event prediction \cite{leblay2018deriving}, question answering \cite{jia2018tequila}, and recommendation systems \cite{trivedi2019dyrep}. Different from static knowledge graphs (KGs), each fact in TKGs additionally associates with temporal information and is represented in the form of ($subject$ $entity$, $relation$, $object$ $entity$, $timestamp$), where $timestamp$ in most temporal reasoning tasks is sparsity.

Reasoning tasks on TKGs contain the entity reasoning task and the relation reasoning task.
The former predicts the missing entity under a specific entity and relation, i.e. ($subject$ $entity$, $relation$, $?$, $timestamp$). Similarly, the latter is represented as ($subject$ $entity$, $?$, $object$ $entity$, $timestamp$).
Additionally, according to the occurrence time of missing facts, temporal reasoning tasks are categorized into two subtasks, $extrapolation$ \cite{trivedi2017know} and $interpolation$ \cite{goel2020diachronic}.
Given training facts within the interval [$t_I$, $t_T$], $interpolation$ predicts missing facts that happened in [$t_I$, $t_T$], while $extrapolation$ makes predictions for future missing facts $t$$>$$t_T$. Therefore, $extrapolation$ is more challenging because there is no real-time information for reference.
This paper mainly focuses on the entity reasoning $extrapolation$ task.

\begin{figure}[t]
	\begin{center}
		
		\includegraphics[width=1\linewidth]{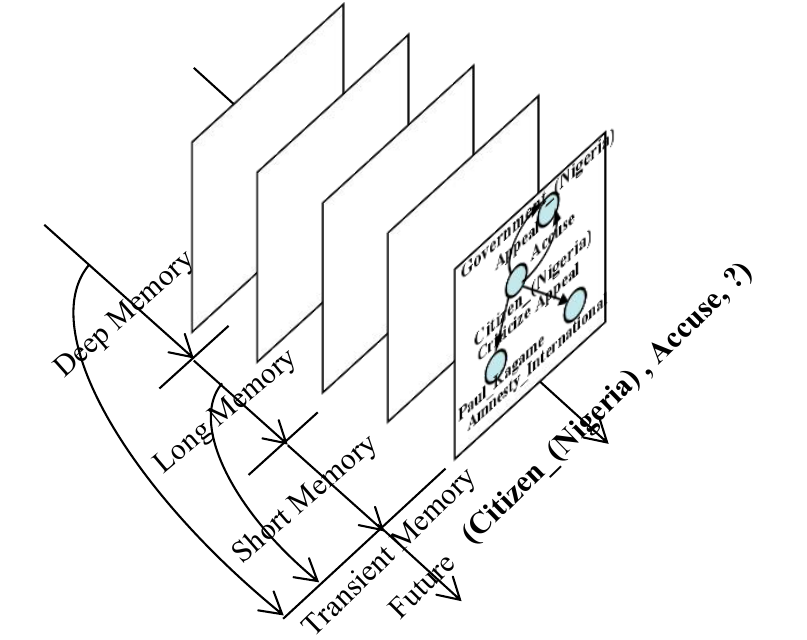}
	\end{center}
	\caption{An example for entity reasoning task over TKGs. Deep memory, long-term memory, short-term memory, and transient memory follow a chronological order.}
	\label{fig2}
\end{figure}

Recent $extrapolation$ methods \cite{tao2021temporal} realize excellent prediction performance. For instance, the recurrent event network (RE-Net) \cite{jin2019recurrent} extracted temporally adjacent history facts for the given entity as a sequential to predict future events. The temporal copy-generation network (CyGNet) \cite{zhu2020learning} mainly focused on learning repetitive patterns from historical facts. Considering RE-Net and CyGNet extracted information from pre-assigned history facts and neglected the structural dependencies, the recurrent evolutional graph convolution network (RE-GCN) \cite{li2021temporal} adopted all temporally adjacent facts to extract sequential patterns by the evolutional representation method. For the $extrapolation$ task, existing methods have the following restrictions:
{\bfseries (1)} Neglecting the impact of time dependence on the reasoning performance;
{\bfseries (2)} Ignoring the associations between transient memories, long-short-term memories, and deep memories;
{\bfseries (3)} Multi-hop aggregation in the structural encoder affects the earlier aggregated information;
{\bfseries (4)} No clear events dissolution constraint.

Specifically, recent facts show greater influences on future predictions than earlier facts \cite{jin2019recurrent}. Namely, long-term memories are more likely to ignore recent changes in target entities. However, transient memories are difficult to dive deeply into historical facts. 
Therefore, to accurately predict the missing entities in the future, it is necessary to adaptively understand the correlation between the missing entities and historical facts.
We annotate restrictions {\bfseries (1-2)} as a unified question - how to incorporate various memories and jointly make predictions.
Another question exists in extracting multi-relational information.
The multi-relational graph aggregator \cite{hataya2021graph}, which can aggregate information from multi-relational and multi-hop neighbors, is applied in many TKGs reasoning methods \cite{jin2019recurrent, li2021temporal}. 
However, the aggregation operation for multi-hop neighbors will gradually destroy the earlier aggregated information.
Additionally, previous methods pay less attention to the dissolution of facts, such as `$sb$ $died$ $in$ $placeA$, $timestamp$'.
In fact, the dissolved facts should not be considered again in future predictions, i.e. `$who$ $died$ $in$ $placeA$, $>timestamp$'.

Considering these problems, we propose a {\bfseries M}emory-{\bfseries T}riggered {\bfseries D}ecision-{\bfseries M}aking network (MTDM) and introduce the dissolution learning constraint for training the network, which utilizes deep memories, long-short term memories, and transient memories, as shown in Fig. \ref{fig2}. 
MTDM consists of a transient learning network that simulates unconditional reflex and a time-aware recurrent evolution network that simulates conditioned reflex. For instance, the central nervous system gives instructions after the knee responds when it is struck.
We treat the transient learning network as a static knowledge graph reasoning task and the time-aware recurrent evolutional network as a sequence of recurrent evolution units.
The recurrent evolution unit gradually updates the attribute representations of entities based on the initial entity representations.
Assume that entity representations are static and attributes of entities are dynamic. 
For instance, given a series of triplets,
($Albert$ $Einstein$, $BornIn$, $German$ $Empire$) $\rightarrow$
($Albert$ $Einstein$, $GraduateFrom$, $University$ $of$ $Zurich$) $\rightarrow$
($Albert$ $Einstein$, $ExpertIn$, $Physics$) $\rightarrow$
($Albert$ $Einstein$, $DiedIn$, $US$),
the initial embedding of $Albert$ $Einstein$ denotes the static entity representations, and $Getting$ $old$ is a kind of dynamic attribute.

The transient learning network and time-aware recurrent evolutional network jointly encode historical facts into entity attribute representations to make predictions. 
Since the problem of excessive information coverage during the multi-hop aggregation process, the structural encoder is implemented by a novel crafted {\bfseries Res}idual {\bfseries GCN} (Res-GCN).
Additionally, the evolution unit still includes a time encoder and a reset gating unit.
Since the gated recurrent unit (GRU) exhibits powerful analytical capabilities in fields like dialogue generation \cite{tran2017semantic} and machine translation \cite{zhang2020neural}, we use GRU as the time encoder of the evolution unit to record beneficial long-term information.  
To break through the time constraints in the reasoning process and automatically determine the history length, we introduce a reset gating unit to aggregate long-short-term memories and deep memories.
Next, the control gating unit that selects the evolutional entity representations or transient entity representations is adopted.
Finally, we introduce the dissolution learning constraint by constructing dissolved facts as negative samples to enhance the network reasoning performance.

In summary, this paper has the following contributions:
(1) We propose a {\bfseries M}emory-{\bfseries T}riggered {\bfseries D}ecision-{\bfseries M}aking network (MTDM). MTDM combines transient memories, long-short-term memories, and deep memories, which alleviates the time dependence problem in temporal reasoning tasks.
(2) We introduce a dissolution learning constraint for MTDM, which can improves the cognitive ability of MTDM on facts dissolution.
(3) The combination of various memories accelerates the model convergence speed, e.g. MTDM reaches the optimal performance in the 8th epoch on the ICEWS14 dataset \cite{trivedi2017know}, while the best available work converges in the 25th epoch. Therefore, MTDM enables up to 2.85 times the training speedup compared to the most advanced model.
(4) MTDM achieves state-of-the-art performance on several standard TKGs benchmarks, including WIKI \cite{leblay2018deriving}, YAGO \cite{mahdisoltani2014yago3}, ICEWS14 \cite{trivedi2017know}. 

\begin{figure*}[t]
	\begin{center}
		
		\includegraphics[width=0.9\linewidth]{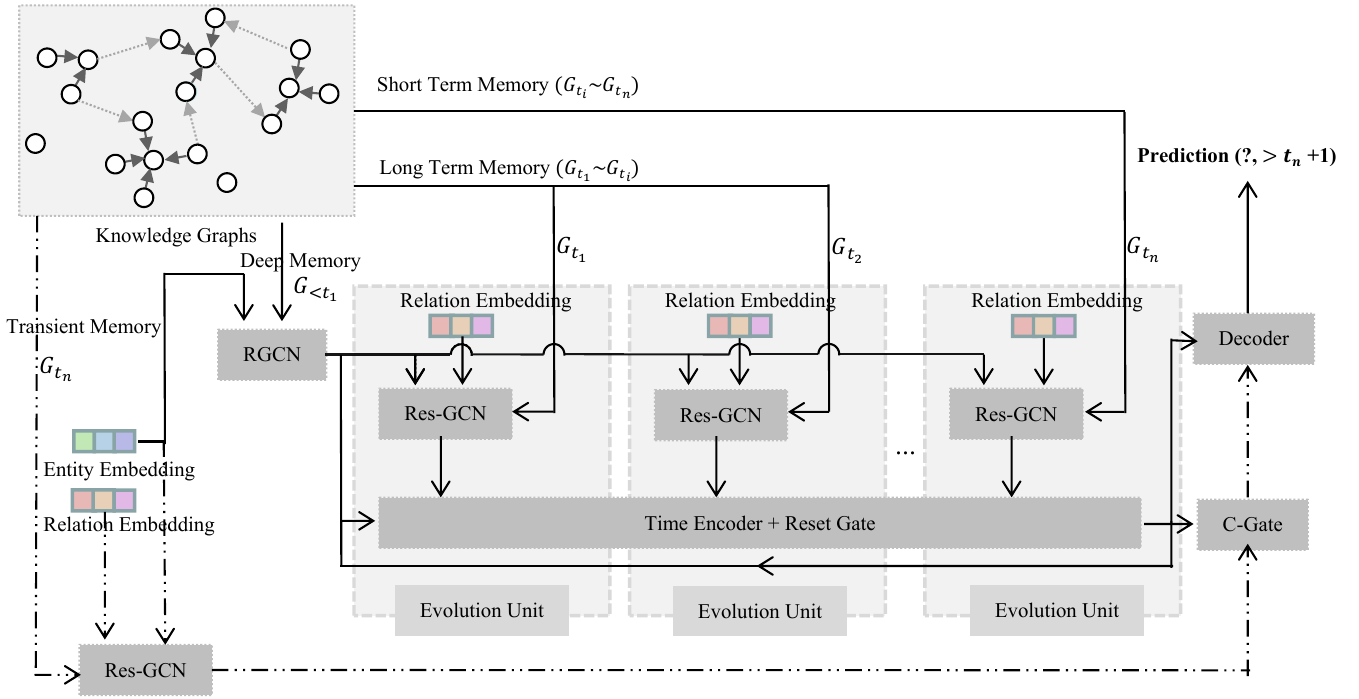}
	\end{center}
	\caption{The proposed Memory-Triggered Decision-Making Network (MTDM).}
	\label{fig1}
\end{figure*}

\section{Related work}
A knowledge graph is a multi-relational graph, which consists of entities, relations, and attributes descriptions \cite{ji2021survey}. 
\subsection{Reasoning based on Static knowledge graphs (KGs)}
Static knowledge graphs ignore the sequence of events.
Recent researches in knowledge reasoning based on KGs (KGs reasoning) focus on knowledge representation learning, namely, mapping entities and relations into low-dimensional vectors.
TransE \cite{bordes2013translating} proposed to embed entities and relationships of multi-relational data into low-dimensional vector spaces.
TransH \cite{wang2014knowledge} took relations as a hyperplane, which aimed to incorporate mapping properties between relations.
TransR \cite{lin2015learning} built entity and relation representations in separate spaces and then learn embeddings by projecting entities from entity space to corresponding relation space.
ComplEx \cite{trouillon2016complex} introduced complex value embedding, which can handle various binary relationships, including symmetric and antisymmetric relations.
RotatE \cite{sun2019rotate} realized the KG embedding by relational rotation in complex space and defined each relation as a rotation from the source entity to the target entity in the complex vector space.
DistMult \cite{yang2014embedding} proved that generalized bilinear formulation achieved excellent link prediction performance.
Researchers also adopted learned relation representations to learn logical rules.
SimplE \cite{kazemi2018simple} proposed a simple enhancement of Canonical Polyadic (CP) decomposition to learn embeddings of entities dependently and the complexity grows linearly with the size of embeddings.
ConvE \cite{dettmers2018convolutional} used neural convolution to build the interactions between entities and relations after mapping entities and relations into variables.
RSN \cite{guo2019learning} was proposed to exploit long-term relational dependencies in knowledge graphs, which employed a skipping mechanism to bridge the gaps between entities.
The designed recurrent skip mechanism helped distinguish relations and entities.

Moreover, researchers \cite{yao2019kg} applied the transformer into the KGs completion and named their framework as knowledge graph bidirectional encoder representations from the transformer (KG-BERT). KG-BERT treated facts in KGs as a textual sequence.
The structure-aware convolutional network (SACN) \cite{shang2019end} combined the benefit of the graph convolution network (GCN) \cite{zhang2019graph} and ConvE, which consisted of a weighted graph convolutional network as an encoder and a convolutional network called Conv-TransE as a decoder.

\subsection{Reasoning based on Temporal knowledge graphs (TKGs)}
Temporal knowledge graphs gradually attract the attention of researchers for the structured knowledge existing within a specific
period and the evolution of facts following the time order.

Researchers of \cite{leblay2018deriving} considered the temporal knowledge graphs by annotating relations between entities within time intervals.
HyTE \cite{dasgupta2018hyte} projected entities and relations into a hyperplane that associated with the corresponding timestamp. 
\cite{jiang2016encoding} introduced a temporal order constraint based on a manifold regularization term considering the happen time of facts.
Han et.al \cite{han2020graph} introduced the Hawkes process into the TKG reasoning task.
Sankalp Garg et.al \cite{garg2020temporal} proposed the temporal attribute prediction, in which each entity extra contained the attribute description.
RE-Net \cite{jin2019recurrent} was an autoregressive architecture for predicting future missing facts. History facts that share the same subject entity with the predictive event are selected as past knowledge.
Tao et.al \cite{tao2021temporal} proposed a TKGs reasoning method based on the reinforcement learning technique.
Researchers trained a time-aware agent to navigate the graph conditioned on known facts to find predictive routes.
CluSTeR \cite{li2021search} predicted future missing facts in a two-stage manner, Clue Searching and Temporal Reasoning.
The clue searching stage induced clues from historical facts based on reinforcement learning. The temporal reasoning stage deduced answers from clues using a graph convolution network. 
RE-GCN \cite{li2021temporal} learned the evolutional representations of entities and relations at each timestamp recurrently.
All history facts are adopted to construct a knowledge graph to learn structural dependencies.

Compared with these temporal reasoning methods, this work also combines deep memories and transient memories by constructing a joint prediction network for solving the time dependence of the reasoning process under the supervision of the dissolution learning constraint.

\section{Problem Formulation}
We consider a static knowledge graph $\mathcal{S}$ as a multi-relational graph, $\mathcal{S}$ = \{$\mathcal{E}$, $\mathcal{R}$, $\mathcal{F}$\}, where $\mathcal{E}$, $\mathcal{R}$, $\mathcal{F}$ are sets of entities, relations and facts. Each fact in the graph can be represented in the form of triples ($s$, $r$, $o$), where $s$ denotes subject entities, $o$ denotes object entities, and $r$ is relations between subject entities and object entities. Then, a TKG $\mathcal{G}_{t_1\thicksim t_n}$ is formalized as a sequence of snapshots $\mathcal{S}$ within a temporal interval [$t_1$, $t_n$], $\mathcal{G}_{t_1\thicksim t_n} = \{t_1: \mathcal{S}_{t_1}, t_{2}: \mathcal{S}_{t_1}, \cdots, t_n: \mathcal{S}_{t_n}\}$, where $t_n$- $t_1$ denotes the time length of selected history facts. $\mathcal{G}_{t_1\thicksim t_n}$ represents the $n$ historical snapshots closest to the predictive timestamp $t_p$. For convenience, we provide symbolic descriptions of time intervals in Table \ref{tab6}. Our goal is to predict events that happened in the future $t_p$, including the object entity prediction, ($s$, $r$, ?, $t_p > t_n$), and subject entity prediction, (?, $r$, $o$, $t_p > t_n$).
Towards this end, there are two different reasoning modes. 
The first one \cite{li2021temporal} directly utilizes the known facts as recent historical facts to predict the future missing facts.
The other \cite{jin2019recurrent} first generates possible events that happened in the future based on known facts and then makes predictions for the future by using these generated events as historical facts.
This work adopts the first reasoning mode.

Taking the object entity prediction as an example, given the constructed TKGs by known historical facts, subject entities $s$ and specific relations $r$ in predictive timestamp $t_p$, the problem of the temporal reasoning is solved by:
\begin{equation}
\max\vec{p}\left(o_{t_p} \mid s_{t_p}, r_{t_p},\mathcal{G}_{\le t_n}, \mathbf{H}_{init}, \mathbf{R}, \theta \right)
\end{equation}
where $\mathbf{H}_{init}$ denotes the initial entity embeddings, $\mathbf{R}$ describes the initial relation embeddings, and $\theta$ represents all network parameters.
Strictly speaking, the relation embeddings show little relevance to time. For instance, ``$Barack$ $obama$ $was$ $born$ $in$ $Hawaii$, $USA$", where `$was$ $born$ $in$' denotes a specific meaning regardless of when the fact happens. 
Therefore, we structure the relation representations as unified static embeddings.

\renewcommand{\arraystretch}{1.3} 
\begin{table}[tp]

	\centering
	
	\fontsize{8}{8}\selectfont
	
	\begin{threeparttable}
		
		\caption{Symbolic descriptions of time intervals.}
		
		\label{tab6}
		
		\begin{tabular}{c|ccc}
			
			\toprule
			
			Time Interval &\multicolumn{3}{c}{Descriptions}\cr
			
			\midrule
			
			$t_p$ & \multicolumn{3}{c}{The predictive timestamp.}\cr
			[$t_I$, $t_T$] & \multicolumn{3}{c}{The time interval of the training data.}\cr
			[$t_T$, $t_E$] & \multicolumn{3}{c}{The time interval of the evaluation data.}\cr
			[$t_1$, $t_n$] & \multicolumn{3}{c}{Recent historical facts.}\cr
			
			\bottomrule
			
		\end{tabular}
		
	\end{threeparttable}
	
\end{table}

\section{Overview of MTDM}\label{sec:method}

MTDM combines a {\bfseries T}ransient {\bfseries L}earning {\bfseries N}etwork (TLN) and a {\bfseries T}ime-aware {\bfseries R}ecurrent {\bfseries E}volutional {\bfseries N}etwork (TREN). TLN makes predictions towards the missing entities by learning knowledge from transient memories. Then, TREN concludes the entity attribute representations by incorporating new knowledge following the chronological order, from the deep memory, the long-term memory to the short-term memory.
In this work, all facts within a temporal interval (or in a timestamp) constitute a snapshot $\mathcal{S}$ of TKGs $\mathcal{G}$. Given recent historical snapshots, TLN uses the multi-residual aggregator to extract the attribute representations of entities by mapping each entity to low-dimensional embeddings. 
TREN consists of a sequence of evolution units, which recursively updates the entity attribute representations based on the initial entity embeddings $\mathbf{H}_{init}$. 
Next, the entity attribute representations extracted by the TLN and TREN are selected through the control gating unit for decision-making. 

Figure \ref{fig1} describes the overall framework of the MTDM. The core sight is that the input of the encoder links the transient memories, long-short-term memories, and deep memories of TKGs. 
The input of the decoder depends on the control gating unit of TLN. When TLN performs non-ignorable prediction errors, only entity representations of TREN participate in the final decision.

\subsection{The time-aware recurrent evolutional network, TREN}\label{remode}
We first introduce TREN since the final outputs of it are used in TLN.

TREN can be regarded as a sequence of evolution units.
Although recent historical facts contribute more than earlier historical facts when reasoning future events, long-term memories still play an indispensable role.
To extract as much historical information as possible, TREN incorporates all facts that occurred in history.

\subsubsection{Deep memory}
Considering limited computer resources and computing efficiency, evolution networks only incorporate all recent historical facts within a period ($t\in[t_1, t_n]$), which means it is difficult to extract information beyond the coverage interval ($t< t_1$). To take advantage of historical facts, we apply all historical facts not covered by TREN to construct a static knowledge graph, which we call deep memories. 
Then, RGCN \cite{schlichtkrull2018modeling} is adopted to extract entity attribute representations $\mathbf{H}_{s,<t_1}$ that used as the initial embeddings of TREN ($\vec{e}_{s, t_1-1}$) from deep memories. 
Formally, the RGCN aggregator is defined as follows:
\begin{equation}
\vec{e}_{s}^{(l+1)}=f\left( \sum_{(r, o), \exists(s, r, o) \in \mathcal{F}} \left(\frac{\mathbf{W}_{r}^{(l)} \vec{e}_{{o}}^{(l)}}{|\mathcal{N}_s|} \right)+\mathbf{W}_{loop}^{(l)} \vec{e}_{s}^{(l)}\right)
\end{equation}
where $\mathbf{W}_{r}^{(l)}$ denotes the learnable relation embedding matrix of the $l$-layer, and $\vec{e}^0$ = $\mathbf{H}_{init}$.
In this way, TREN can extract helpful information ($\vec{e}_{s, t_1-1}$ = $\vec{e}_{s}^{w}$) from deep memories that have never been covered.

\subsubsection{The structural encoder}
Next, we propose a novel structural encoder Res-GCN to extract structural information from long-short-term memories.
The structural encoder aims to capture the association between entities through concurrent facts, which generates entity attribute representations based on recent historical snapshots $\mathcal{G}_{t_1\thicksim t_n}$. Considering the multi-hop aggregation in relation-aware GCN \cite{li2021temporal} is less sensitive to the earlier aggregated information, an $w$-layer Res-GCN is built to model the entity attribute representations.
Specifically, given the set of all known historical facts $\mathcal{F}_{t}$ at timestamp $t$, the entity attribute representations $\vec{e}_{s, t}^{(l)}$ of $l$-hop aggression layer ($l\in[1,w]$) aggregate information from its historical entity attribute representations $\vec{e}_{o, t}^{(l - 1)}$.
\begin{equation}
\vec{e}_{s, t}^{loop}=\mathbf{W}_{loop} \vec{e}_{s, t}^{(0)}
\end{equation}
\begin{equation}
\vec{res}_{s, t}^{(l+1)}=\left(\frac{1}{|\mathcal{N}_s|} \sum_{(r, o), \exists(s, r, o) \in \mathcal{F}_{t}} \left(\mathbf{~W}_{e}\cdot\vec{e}_{o, t}^{(l)}+\mathbf{~W}_{r}\cdot\vec{r}\right)\right)
\end{equation}
\begin{equation}
\vec{e}_{s, t}^{(l+1)}=f\left(\vec{e}_{s, t}^{loop} + \sum_{i=0}^{l+1} \vec{res}_{s, t}^{(i)} \right)
\end{equation}
where $\mathbf{W}_{loop}$ denotes the self-loop embedding matrix of entities, $\vec{e}_{s, t}^{(0)}$ is the initial attribute representations in timestamp $t$, $\mathbf{~W}_{e}$ and $\mathbf{~W}_{r}$ aim to map entity attribute representations and relation representations to the same space. Subject entities and object entities share the same entity embedding matrix. $\mathcal{N}_s$ represents the set of neighboring entities of $s$, and its size $|\mathcal{N}_s|$ serves as a normalization constant for averaging neighborhood information. $\vec{res}_{s, t}^{(l)}$ describes the residual attribute representations learned by $l$-hop aggregation and $f$ means activation function.

For convenience, the formula description of the extraction process for the structural knowledge is,
\begin{equation}\label{eq2}
\vec{e}_{s, t}^{(w)} = \text{Res-GCN}(\vec{e}_{s, t_1-1}, \mathcal{F}_{t})
\end{equation}
Here $\vec{e}_{s, t_1-1}$ is the initial entity attribute representations, and $\mathcal{F}_{t}$ denotes known historical facts.

\subsubsection{The time encoder \& reset gate}
Finally, the time encoder recursively updates the entity attribute representations by combining the extracted aggregation information [$\vec{e}_{s, t_1}^{(w)}, \vec{e}_{s, t_2}^{(w)}, \cdots, \vec{e}_{s, t_n}^{(w)}$].
Compared with TLN, TREN behaves more sensitive to temporal information and is prone to over-smoothing. Therefore, we use GRU as the time encoder in TREN.
The input of the time encoder at timestamp $t$ includes the aggregate information $\vec{e}_{s, t-1}^{(w)}$ extracted by the structural encoder and the GRU output $\mathbf{H}_{s,t-1}^{\text{TREN}}$ at timestamp $t-1$.
\begin{equation}
\mathbf{H}_{s,t}^{\text{TREN}}=\text{GRU}\left(\vec{e}_{s, t-1}^{(w)}, \mathbf{H}_{s,t-1}^{\text{TREN}}\right)
\label{eq1}
\end{equation}
After that, to automatically determine the history length and effectively retain deep memories, we introduce a reset gating unit after each time encoder.
\begin{equation}
\mathrm{V}_{t}=\sigma\left(\mathbf{W}_{\mathrm{V}}\cdot \mathbf{H}_{s,t}^{\text{TREN}}+\mathrm{b}_{\mathrm{V}}\right)
\end{equation}
\begin{equation}
\mathbf{H}_{s,t}^{\text{TREN}}=\mathrm{V}_{t} \cdot \mathbf{H}_{s,t}^{\text{TREN}} +\left(1-\mathrm{V}_{t}\right) \cdot \vec{e}_{s, t_1-1}
\end{equation}
The TREN will automatically neglect previous historical facts once the model assigns initial values to $\mathrm{V}_{t}$, $\mathrm{V}_{t}$ = 0.
After the $n$-layer time encoders and gating units, we obtain the final long-term entity attribute representations $\mathbf{H}_{s,t_n+1}^{\text{TREN}}$.

\subsection{The transient learning network, TLN}
TLN consists of a structural encoder Res-GCN and a time encoder that selects entity attribute representations - the control gating unit.
\subsubsection{The structural encoder}
Similar to TREN, TLN also adopts the Res-GCN as the structural encoder.
\begin{equation}
\vec{h}_{s, t}^{(w)} = \text{Res-GCN}(\mathbf{H}_{init}, \mathcal{F}_{t})
\end{equation}
We use the final multi-hop aggregate information $\vec{h}_{s, t_n}^{(w)}$ as attribute representations of entities $s$ at timestamp $t$ = $t_n$.
\subsubsection{The Time encoder}
As we have mentioned, not all representations extracted by TLN and TREN can efficiently make predictions about missing facts.
Thus, we introduce a control gating unit to dynamic select the attribute representations, as follows,
\begin{equation}
\mathbf{H}_{s,t_n+1}^{\text{TLN}}=\mathrm{U}_{t_n} \cdot \vec{h}_{s, t_n}^{(w)} +\left(1-\mathrm{U}_{t_n}\right) \cdot \mathbf{H}_{s,t_n+1}^{\text{TREN}}
\end{equation}
\begin{equation}
\mathrm{U}_{t_n}=\sigma\left(\mathbf{W}_{\mathrm{U}}\cdot \vec{h}_{s, t_n}^{(w)}+\mathrm{b}_{\mathrm{U}}\right)
\end{equation}
where $\mathbf{H}_{s,t_n+1}^{\text{TLN}}$ and $\mathbf{H}_{s,t_n +1}^{\text{TREN}}$ represent the final attribute representations of $s$ extracted from the TLN and TREN for predicting the missing facts in $t_p$, respectively.
In the training process, $t_p$ = $t_{n+1}$. However, in the testing process, $t_p\gg$ $t_{n+1}$.
The time gate $\mathrm{U}_{t_n}$ is determined by the $\vec{h}_{s, t_n}^{(w)}$.
Namely, $\mathbf{H}_{s,t_n+1}^{\text{TREN}}$ replaces $\vec{h}_{s, t_n}^{(w)}$ as the attribute representations of entities $s$ once transient memories seviously affect the prediction performance, $\forall u \in \mathrm{U}_{t_n}$, $u$ in $\{0, 1\}$.

\subsection{The Decoder}

Previous works \cite{malaviya2020commonsense, vashishth2020interacte} show that KG reasoning tasks with the convolutional neural network as the decoder displays excellent performance. Thus, we choose the ConvTransE as the network decoder following \cite{shang2019end}, which contains several convolution layers, and a fully connected layer. For reasoning missing entity facts ($t_p> t_n$), both entity attribute representations ($\mathbf{H}_{t_n+1}^{\text{TREN}}, \mathbf{H}_{t_n+1}^{\text{TLN}}$) extracted by the MTDM and static relation representations $\mathbf{R}$ are used as decoder inputs.
Thus, the entity prediction probability is defined as,
\begin{equation}
\mathbf{H}_{t_p} = <\mathbf{H}_{t_n+1}^{\text{TREN}}, \mathbf{H}_{t_n+1}^{\text{TLN}}>
\end{equation}
\begin{equation}
\mathbf{P}_{o,t_p} = \operatorname{ConvTransE}\left(\mathbf{H}_{s_{t_p},t_n+1}^{\text{TREN}}, \mathbf{H}_{s_{t_p},t_n+1}^{\text{TLN}}, \mathbf{R}(r_{t_p})\right)
\end{equation}
\begin{equation}
\vec{p}_{t_p}(o) = \vec{p}\left(o \mid s_{t_p}, r_{t_p}, \mathcal{G}_{<t_p}\right)=\sigma\left(\mathbf{P}_{o,t_p}\otimes\mathbf{H}_{t_p}^\dagger \right)
\end{equation}
where $\mathbf{H}_{t_n+1}^{\text{TREN}}$, $\mathbf{H}_{t_n+1}^{\text{TLN}}$ are sets of $\mathbf{H}_{s, t_n+1}^{\text{TREN}}$, $\mathbf{H}_{s, t_n+1}^{\text{TLN}}$, $<$,$>$ denotes the matrix concatenate, $\mathbf{P}_{o,t_p}$ is the entity prediction vector decoded by the ConvTransE, $\otimes$ denotes the matrix multiplication, $\dagger$ means matrix transpose, and $\sigma(\cdot)$ is the sigmoid function.
Since the static relation representations significantly affect the entity reasoning performance, we construct the relation prediction probability for better extracting relation representations, which is expressed as,
\begin{equation}
\mathbf{P}_{r} = \operatorname{ConvTransE}\left(\mathbf{H}_{s_{t_p},t_p},\mathbf{H}_{o_{t_p},t_p}\right)
\end{equation}
\begin{equation}
\vec{p}_{t_p}(r) = \vec{p}\left(r \mid s_{t_p}, o_{t_p}, \mathcal{G}_{<t_p}\right)=\sigma\left(\mathbf{P}_{r}\otimes\mathbf{R}^\dagger \right)
\end{equation}

\subsection{Parameter Learning}
Based on pediction probabilities $\vec{p}_{t_p}(o), \vec{p}_{t_p}(r)$, the loss function for entity prediction is defined as,
\begin{equation}
\mathcal{L}_e=-\sum \left((1-\lambda_{1})\cdot\log\vec{p}_{t_p}(o_{t_p}) +\lambda_{1}\cdot \log \vec{p}_{t_p}(r_{t_p})\right)
\end{equation}
where ${(\mathrm{s}_{t_p}, \mathrm{r}_{t_p}, \mathrm{o}_{t_p}) \in \mathcal{F}_{t_p}}$. We experimentally assign the value for the hyper-parameter $\lambda_{1}$.

To reflect the deep memories in learned entity attribute representations, following \cite{li2021temporal}, we constraint the difference between the deep memory representations $\mathbf{H}_{<t_1}$ and the entity attribute representations $\mathbf{H}_{t}^{\text{TREN}}$ of the same entity to not exceed the threshold $\theta_{t}$. 
Additionally, the embedding threshold will increase over time since recent historical facts gradually modify the deep memory.
\begin{equation}\label{angle}
\mathcal{L}_{d} =\sum_{t=t_1}^{t_n} \max \left\{\cos \theta_{t} -\cos \left(\mathbf{H}_{<t_1}, \mathbf{H}_{t}^{\text{TREN}}\right), 0\right\}
\end{equation}
where $\theta_{t}$ = $\gamma\cdot (t- t_1 + 1)$, $\gamma$ represents the angle stride.

We have experimentally found that transient memories greatly affect long-term memory-dependent decisions. Therefore, we introduce a 0-1 constraint on gating parameters $\mathrm{U}_{t}$.
\begin{equation}
\mathcal{L}_g= \bar{\mathrm{U}_{t}}\cdot(1- \bar{\mathrm{U}_{t}})
\end{equation}
where $\bar{\mathrm{U}_{t}}$ denotes the average value of gating parameters.

\subsection{Dissolution Learning}

By observing training data in TKGs reasoning tasks, we found that most relations between two observed entities do not change, while only a few new relations have formatted and dissolved since the last observation. 
Thus, we construct dissolved events for each prediction timestamp.
Specifically, for predicted facts ($s$, $r$, ?, $t_p$), $t_p > t_n$, we collect historical facts $adv = (s, r, ?)$ within a temporal interval [$t_1, t_n$], $adv \in\mathcal{F}_{t_1\thicksim t_n}$, $adv \not\in\mathcal{F}_{t_p}$.
Then we assign temporal information for these newly constructed facts, \{$adv$, $t_p$\}$\in\mathcal{F}_{adv}$.
\begin{equation}\label{eq5}
\mathcal{L}_{adv}=  \sum_{(\mathrm{s}, \mathrm{r}, \mathrm{o}) \in \mathcal{F}_{adv}}\log\vec{p}_{t_p}(o = o_{t_p})
\end{equation}

Based on above mentioned loss functions, the final loss constraint is defined as:
\begin{equation}\label{eq4}
\mathcal{L}= \mathcal{L}_e + \lambda_{2}\cdot\mathcal{L}_d + \lambda_{3}\cdot\mathcal{L}_g + \lambda_{4}\cdot\mathcal{L}_{adv}
\end{equation}
$\lambda_{1-4}$ are hyper-parameters that distinguish the importance of different tasks. $\lambda_{4}$ = 0 by default. We ablatively study the impact of dissolution learning constraint $\mathcal{L}_{adv}$ on the reasoning performance in Section \ref{dl}.
The process for entity prediction task based on MTDM is highlighted in Algorithm \ref{example}.

\begin{algorithm}[t]
	\caption{The algorithm of the memory-trigged decision making network.} \label{example}
		\BlankLine
		\KwIn{Subject entities $s$ and specific relations $r$ in predictive timestamp $t_p (t_p> t_n)$, the initial entity embeddings $\mathbf{H}_{s}$, and the initial relation embeddings $\mathbf{R}$.}
		\KwOut{Make predictions for object entities $o$.}
		/* TLN: Calculate attribute representations $\vec{h}_{s, t_n}^{(w)}$ by the structural encoder SE (Res-GCN).  */
		
		$\vec{h}_{s, t_n}^{(w)} = \text{SE}(\mathbf{H}_{init}, \mathbf{R}, \mathcal{F}_{t_n})$;
		
		/* TREN: Calculate $\mathbf{H}_{<t_1}$ based on deep memories $\mathcal{G}_{<t_1}$ and RGCN.  */
		
		$\mathbf{H}_{<t_1}$ = RGCN($\mathbf{H}_{init}$, $\mathcal{G}_{<t_1}$)
		
		/* TREN: Calculate aggregation information [$\vec{e}_{s, t_1}^{(w)}, \vec{e}_{s, t_2}^{(w)}, \cdots, \vec{e}_{s, t_n}^{(w)}$] using SE for all entities $s$. */
		
		\For{t $\in$ [t$_1$, t$_n$]} {
			$\vec{e}_{s, t}^{(w)} = \text{SE}(\vec{e}_{s, t_1-1}, \mathcal{F}_{t})$}
		/* TREN: Updating entity attribute representations $\mathbf{H}_{s, t}^{\text{TREN}}$, based on the aggregation information [$\vec{e}_{s, t_1}^{(w)}, \cdots, \vec{e}_{s, t_n}^{(w)}$], and time encoder GRU. */
		
		\For{t $\in$ [t$_1$, t$_n$]}{
			
			$\mathbf{H}_{s, t}^{\text{TREN}}=\text{GRU}\left(\vec{e}_{s, t-1}^{(w)}, \mathbf{H}_{s, t-1}^{\text{TREN}}\right)$
			
			$\mathbf{H}_{s, t}^{\text{TREN}}=\mathrm{V}_{t} \cdot \mathbf{H}_{s, t}^{\text{TREN}} +\left(1-\mathrm{V}_{t}\right) \cdot \vec{e}_{s, t_1-1}$
		}
		
		/* TLN: The control gating unit selects attribute representations $\mathbf{H}_{s, t}^{\text{TLN}}$.  */
		
		$\mathbf{H}_{s,t_n+1}^{\text{TLN}}=\mathrm{U}_{t_n} \cdot \vec{h}_{s, t_n}^{(w)} +\left(1-\mathrm{U}_{t_n}\right) \cdot \mathbf{H}_{s,t_n+1}^{\text{TREN}}$
		
		/* Calculate the entity prediction probability $\vec{p}_{t_p}(o_{t_p})$ by ConvTransE based on $\mathbf{H}_{s_{t_p},t_n+1}^{\text{TLN}}$ and $\mathbf{H}_{s_{t_p},t_n+1}^{\text{TREN}}$. */
		
		/* Calculate the relation prediction probability $\vec{p}_{t_p}(r_{t_p})$ by ConvTransE based on $\mathbf{H}_{s_{t_p},t_n+1}^{\text{TLN}}$ and $\mathbf{H}_{s_{t_p},t_n+1}^{\text{TREN}}$. */
		
		/* Run network pretraining only one epoch and calculate the value for $\mathrm{U}_{t_n}$. */
		
		$\mathcal{L}= \mathcal{L}_e + \lambda_{2}\cdot\mathcal{L}_d + \lambda_{3}\cdot\mathcal{L}_g$
		
		/* Run network training and update model parameters with the following constraint. */
		
		$\mathcal{L}= \mathcal{L}_e + \lambda_{2}\cdot\mathcal{L}_d$
\end{algorithm}

\section{Experiments}
\subsection{Experimental Setup}
We evaluate the performance of the MTDM on three standard datasets and compare it with recent temporal reasoning models.
Code to reproduce our experimental results will be publicly available in \url{https://github.com/Dlut-lab-zmn/MTDM}.

\subsubsection{\bfseries Datesets} Three kinds of TKG datasets are used in our experiments, including the Integrated Crisis Early Warning System sub-dataset, ICEWS14 \cite{trivedi2017know}, corresponding to the facts in 2014, two datasets with temporally associated facts as ($s$, $r$, $o$, [$t_s$, $t_e$]), where $t_s$ is the starting timestamp and $t_e$ is the ending timestamp, WIKI \cite{leblay2018deriving} and YAGO \cite{mahdisoltani2014yago3}.
We follow \cite{li2021temporal} to divide ICEWS14 into training, validation, and testing sets with a proportion of 80\%, 10\%, and 10\% by timestamps, namely,  (timestamps of the training data) $<$ (timestamps of the validation data) $<$ (timestamps of the testing data).

\subsubsection{\bfseries Evaluation Metrics}  
We employ two widely used metrics to evaluate the model performance on extrapolated link predictions, Mean Reciprocal Rank (MRR) and Hits@ \{1, 3, 10\}.
We report experimental results under the time-aware filtered setting following \cite{sun2021timetraveler, ding2021temporal} that only filters out quadruples with query time $t$.
Namely, instead of removing corrupted facts that appear either in the training, validation, or test sets, we only filter from the list all the events that occur on the predictive timestamp \cite{han2020graph}.
Compared with the raw setting and the temporal filtered setting used in \cite{jin2019recurrent}, the setting of \cite{sun2021timetraveler, ding2021temporal} gets more reasonable ranking scores.
For instance, given an evaluation quadruple ($Barack$ $Obama$, $visit$, $Germany$, $Jan. 18, 2013$) and make predictions for quadruple ($Barack Obama$, $visit$, ?, $Jan. 18, 2013$), ($Barack Obama$, $visit$, $Chian$, $Nov. 15, 2009$) in the training set should not be filtered out from the ranking list since the triplet ($Barack Obama$, $visit$, $China$) is only temporally valid on $Nov. 15, 2009$ instead of $Jan. 18, 2013$.

\subsubsection{\bfseries Experimental Settings}  
All experiments are evaluated on an Nvidia GeForce RTX 3090 GPU.
Adam is adopted for updating parameters with a learning rate of 0.001. We set both embedding dimensions of entities and relations to 200, the lengths of long-short-term memories and transient memories to 10, 1. The number of layers of the structural encoder is set to 2.
For ConvTransE, we following the related settings mentioned in RE-GCN \cite{li2021temporal}.
The hyperparameters $\lambda_{1}$, $\lambda_{2}$, $\lambda_{3}$, $\lambda_{4}$ are set to 0.3, 1, 1, 0.
The initial similarity angle is 10$^\circ$, and increases over timestamp with a stride of 10$^\circ$.
All experimental results are obtained under the same experimental conditions, i.e. no additional auxiliary information.
By default, the MTDM takes all known facts as historical facts than generating new possible events as historical information and does not combine the event dissolution learning module.

\subsubsection{\bfseries Baselines}  
We compare the MTDM model with interpolation reasoning models (TTransE \cite{leblay2018deriving}, TA-DistMult \cite{garcia2018learning}, DE-SimplE \cite{goel2020diachronic}, TNTComplEx \cite{lacroix2020tensor}), and state-of-the-art extrapolation reasoning models, including (CyGNet \cite{zhu2020learning}, ReNet \cite{jin2019recurrent}, TANGO \cite{ding2021temporal}, RE-GCN \cite{li2021temporal}, xERTE \cite{han2020explainable}).

\renewcommand{\arraystretch}{1.2} 
\begin{table*}[tp]

	\centering
	
	\fontsize{6}{7}\selectfont
	
	\begin{threeparttable}
		
		\caption{Quantitative results of the proposed Memory-Triggered Decision-Making Network. His$_i$ describes the available history facts. GT means adopting ground truth facts in the evaluation process. The default performances are evaluated on the time-aware filtered setting in \cite{sun2021timetraveler, ding2021temporal}. {\bfseries Bold} denotes the best results.}
		
		\label{tab1}
		
		\begin{tabular}{ccc cccc cccc cccc}
			
			\toprule
			\toprule
			
			\multicolumn{3}{c}{\multirow{2}{*}{{\bfseries Method}}}&\multicolumn{4}{c}{{\bfseries YAGO}}&
			
			\multicolumn{4}{c}{{\bfseries WIKI}}&\multicolumn{4}{c}{{\bfseries ICEWS14s}}\cr
			
			\cmidrule(lr){4-7}\cmidrule(lr){8-11}\cmidrule(lr){12-15}
			
			&&State&MRR &H@1 &H@3&H@10&MRR&H@1&H@3&H@10&MRR&H@1&H@3&H@10\cr
			
			\midrule 
			
			\multicolumn{2}{c}{TTransE \cite{leblay2018deriving}}&-&31.19&18.12&40.91&51.21&29.27 &21.67&34.43&42.39&13.43&3.14&17.32&34.65 \cr
			\multicolumn{2}{c}{TA-DistMult \cite{garcia2018learning}}&-&54.92&48.15&59.61&66.71&44.53 &39.92&48.73&51.71&26.47&17.09&30.22&45.41 \cr
			\multicolumn{2}{c}{DE-SimplE \cite{goel2020diachronic}}&-&54.91&51.64&57.30&60.17&45.43 &42.60&47.71&49.55&32.67&24.43&35.69&49.11\cr
			\multicolumn{2}{c}{TNTComplEx \cite{lacroix2020tensor}}&-&57.98&52.92&61.33&66.69&45.03&40.04&49.31&52.03&32.12&23.35&36.03&49.13 \cr

			\midrule

			\multicolumn{2}{c}{TANGO TuckER \cite{ding2021temporal}}&-&62.50&58.77&64.73&68.63&51.60 &49.61&52.45&54.87&-&-&-&-\cr
			\multicolumn{2}{c}{TANGO Distmult \cite{ding2021temporal}}&-&63.34&60.04&65.19&68.79&53.04 &51.52&53.84&55.46&-&-&-&-\cr
			
			\multicolumn{2}{c}{\multirow{2}{*}{CyGNet \cite{zhu2020learning}} }&-& 60.27& 55.35& 63.80&68.01&49.05&45.61&51.48&53.88&36.45&26.72&40.84&55.46 \cr
			\multicolumn{2}{c}{}&w. eval&71.66&66.03&74.50&83.33&59.67 &53.74&63.27&69.89&37.22&27.59&41.74&55.96\cr
					
			\multicolumn{2}{c}{\multirow{2}{*}{ReNet \cite{jin2019recurrent}} }&-& 60.89& 55.86& 63.93&69.24&49.11&46.76&50.37&53.06&36.67&27.39&40.77&54.81 \cr
			\multicolumn{2}{c}{}&w. eval&73.68&68.12&77.31&82.80&60.28 &55.68&63.01&68.27&37.82&28.35&42.00&56.03\cr

			\multicolumn{2}{c}{\multirow{3}{*}{RE-GCN \cite{li2021temporal}}}&His$_1$&{74.69}&{70.60}&{77.35}&{81.78}&{62.15}&{59.01}&63.93&67.07&35.47&26.45&39.38&53.07\cr
			&&His$_4$&72.39&67.68&75.30&80.44&62.17 &59.06&{64.15}&{67.12}&35.62&26.63&39.08&53.24 \cr
			&&His$_{10}$&70.14&65.81&72.27&77.96&59.97 &56.21&62.28&66.24&{35.85}&{27.02}&{39.16}&{53.53}\cr

			\multicolumn{2}{c}{{MTDM}}&-&77.66&73.85&80.82&83.79&63.18 &60.24&65.31&67.89&37.12&28.15&41.00&54.52\cr

			\midrule
			\multicolumn{2}{c}{xERTE \cite{han2020explainable}}&\multirow{3}{*}{w. GT}&84.19&80.09&88.02&89.78&71.14 &68.05&76.11&79.01&-&-&-&-\cr
			\multicolumn{2}{c}{RE-GCN \cite{li2021temporal}}&&81.91&78.18&84.19&88.62&78.18 &74.48&81.00&84.12&40.26&30.34&45.33&59.32\cr
			\multicolumn{2}{c}{MTDM}&&{\bf 86.27}&{\bf 83.01}&{\bf 88.67}&{\bf 91.41}&{\bf 80.04} &{\bfseries 76.63}&{\bf 83.13}&{\bf 85.64}&{\bf 41.40}&{\bf 31.49}&{\bf 46.43}&{\bf 60.56}\cr

			\bottomrule
			\bottomrule

		\end{tabular}
		
	\end{threeparttable}
	
\end{table*}

\begin{figure}[t]
	\begin{center}
		
		\includegraphics[width=1.0\linewidth]{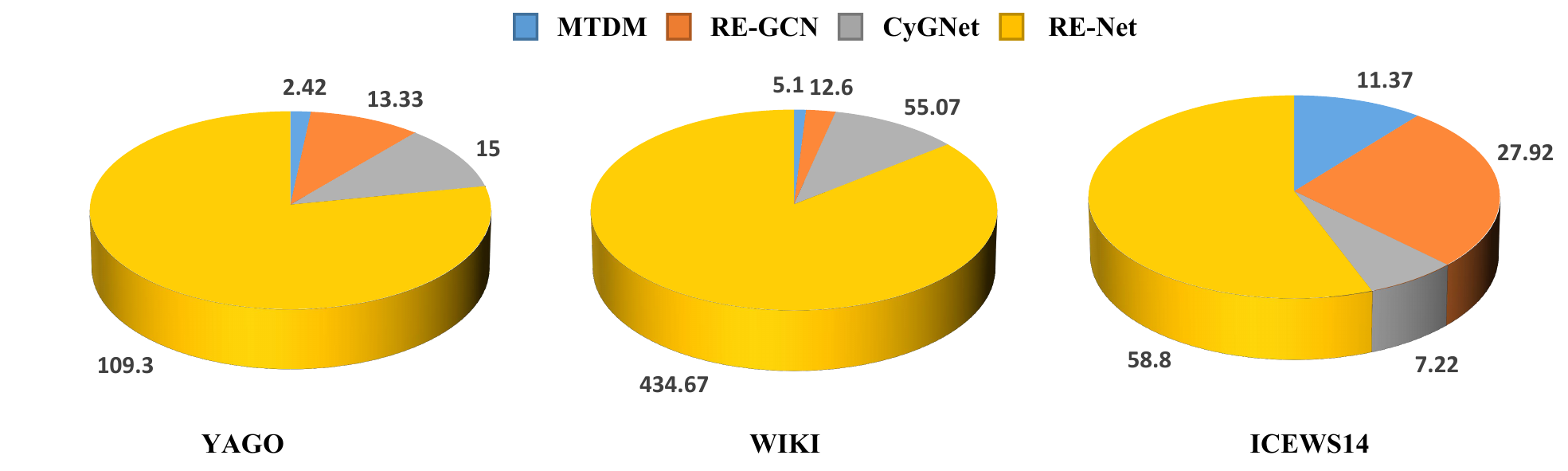}
	\end{center}
	\caption{The comparison of total training time consumption between various TKGs reasoning methods. The $digits$ denote the concrete procedure running minutes.}
	\label{fig3}
\end{figure}
\subsection{Experimental results}
We evaluate the model performance on the time-aware filtered matrix. 
Table \ref{tab1} summarizes entity prediction performance comparison on three benchmark datasets, where $w. eval$ denotes utilizing the evaluation data and training data to train the model. By default, only training data is used in the training process. The training data and evaluation data are adopted as historical facts in the reasoning process.
Specificlly, given triples with timestamps in [$t_I$, $t_T$] as the training set, [$t_T$, $t_E$] as the evaluation set, triples with timestamps in [$t_I$, $t_E$] as historical evaluation facts, predicting triples with the timestamp $t_p >t_{E}$.

Results in Table \ref{tab1} demonstrate that MTDM outperforms all interpolation baselines on three standard datasets, which illustrates the effectiveness of utilizing evolutionary representation learning on entity predictions. 
Additionally, MTDM achieves state-of-the-art performance than all extrapolation baselines.
It is worth noting that only xERTE, RE-GCN, MTDM can tackle unseen entities by dynamically updating new entity representations based on temporal message aggregation, namely, incorporating evaluation data as historical facts to predict future missing entities but not using the evaluation data to train the model. 
Therefore, we also provide the prediction performance with or without utilizing evaluation data when training the model for methods CyGNet, RE-Net.
As we can see, MTDM performs better than CyGNet and RE-Net even these methods combine the evaluation data in the training process.
For instance, the MRR of MTDM is 5.1\% and 7.7\% higher than RE-Net and CyGNet under the YAGO dataset, 4.6\% and 5.6\% higher than RE-Net and CyGNet under the WIKI dataset ($w. eval$).
For clarifying the impact of history length on the prediction performance, three history lengths are set for RE-GCN \cite{li2021temporal}, 1, 4, 10.
We observe that RE-GCN reaches optimal performance for YAGO, WIKI, and ICEWS14 when we set the history length to 1, 4, and 10.
In comparison, MTDM leverages deep memories, long-short-term memories, and transient memories, which alleviates the history dependence on history length.
Thus, MTDM sets the fixed history length for convenience, 10 for TREN, and 1 for TLN.
Even so, MTDM performs better prediction performance than the best RE-GCN, i.e. the MRR of MTDM achieves 3.8\% performance improvement than RE-GCN on the YAGO dataset, 1.6\% improvement on WIKI, and 3.4\% improvement on ICEWS14.

We can draw the following conclusions through the experimental results.
For WIKI and YAGO, short-term memories perform better than long-term memories in predicting future missing entities, i.e. the MRR of RE-GCN under the latest knowledge graph as historical facts achieve 74.69\% prediction accuracy, 6.1\% performance improvement compared with lengths of history to 10 on the YAGO dataset. For ICEWS14, short-term memories combined with long-term memories perform better than only using short-term memories. 
Therefore, MTDM combines the entity attribute representations in TLN selected by the gating unit C-Gate, and the entity attribute representations in TREN that neglect invalid long-term memories for reasoning all standard benchmarks.

In addition, Table \ref{tab1} also provides the comparative experiment results of MTDM, RE-GCN, xERTE with the ground truth events (GT) as historical evaluation facts, where MTDM realizes a 5.1\% MRR improvement than RE-GCN for the YAGO dataset, 2.3\% improvement for WIKI, and 2.8\% improvement for ICEWS14.
Specifically, GT means given triples with timestamps in [$t_I$, $t_T$] as the training set, triples with timestamps in [$t_I$, $t_d$] as historical evaluation facts, and predicting triples with the timestamp $t_{d+1}$ ($t_d > t_E$). Evaluation with GT as historical evaluation facts can better illustrate the transferability of models, namely, incorporating the newest information for predicting the following possible events.
In summary, the MTDM successfully alleviates the temporal dependence problem and outperforms previous works in the temporal reasoning task.

\subsection{Running speed}
To evaluate the efficiency of MTDM, we compare MTDM with the newest TKG reasoning methods, including RE-GCN, RE-Net, and CyGNet, in terms of total training time under the same setting. 
Comparison results are shown in Fig. \ref{fig3}.
We observe that RE-Net takes amounts of time consumption than MTDM, RE-GCN. That is because RE-Net needs to extract temporally adjacent history facts for each given entity as a sequential to predict future events, while MTDM and RE-GCN utilize the evolutionary representation learning.
Meanwhile, MTDM performs better than CyGNet and RE-GCN in time cost, about 4.1 times faster than CyGNet, and 2.9 times faster than RE-GCN.
On the one hand, MTDM only contains a structural encoder, a time encoder with a reset gating unit, and the deep memory learning module that takes less computational time, which means MTDM trains fast in a single epoch. On the other hand, the memory combination operation accelerates the model convergence speed.
For example, MTDM reaches the optimal performance in the 6th and 8th epoch for YAGO and ICEWS14, while the best available work converges in the 24th and 25th epoch.
\subsection{Abalation study}
This subsection investigates the impact of variations in MTDM on reasoning performance, including memory-related variants, model structure-related variants, and data source-related variants. For ablation studies towards various memories, we conduct three sub-MTDM, without the deep memory - w.o DM, without the time-aware recurrent evolutional network - w.o TREN, and without the transient learning network - w.o TLN. Additionally, we experimentally analyze the impact of reset gating units - w.o R$\_$Gate, different structural encoders, different time encoders, and various recurrent modes towards the model prediction. Finally, we conduct researches towards the event dissolution constraint, w. ED, namely, adding dissolved events as negative samples into the training data. Meanwhile, we experimentally analyze the influence of various hyper-parameters on MTDM.

\subsubsection{\bfseries Memory related variants}

As we have mentioned in section \ref{sec:method}, MTDM incorporates transient memories, long-short-term memories, and deep memories by combining a transient learning network and a time-aware recurrent evolutional network. To evaluate the effectiveness of various memories, we conduct ablation studies by removing deep memories, the TREN that incorporating the long-short term memories, and the TLN that extracting the transient memories. Experimental results are illustrated in Table \ref{tab3}. 
We observe that the default MTDM shows better reasoning performance than MTDM w.o DM in the ICEWS14 dataset, which proves deep memories are beneficial to reasoning events with long-term dependence.
For instance, the triplet (`$Person$', `$specialized$ $in$', `$research$ $direction$') is closely correlated with the triplet (`$Person$', `$works$ $at$', `$Company$ $name$'), while deep memories contain amounts of information like these correlations.

Meanwhile, both the inference accuracies of MTDM on the YAGO and ICEWS14 datasets are affected by TREN. For instance, compared with MTDM, MTDM w.o TREN reduces the prediction accuracy by 2\% on the YAGO dataset and 8.7\% on the ICEWS14 dataset. On the one hand, compared with TLN, TREN can excavate the structural information hidden in long-term memories. This structural knowledge contributes a lot in reasoning long-term dependencies in TKGs. On the other hand, the evolutional representations extracted from TREN are helpful for reasoning short-term dependence facts, which are different from representations extracted from transient memories.

Similarly, the prediction performance of MTDM on the YAGO dataset outperforms that of MTDM w.o TLN, which demonstrates the effectiveness of TLN on short-term TKGs prediction. 
\begin{figure}[t]
	\begin{center}
		
		\includegraphics[width=0.85\linewidth]{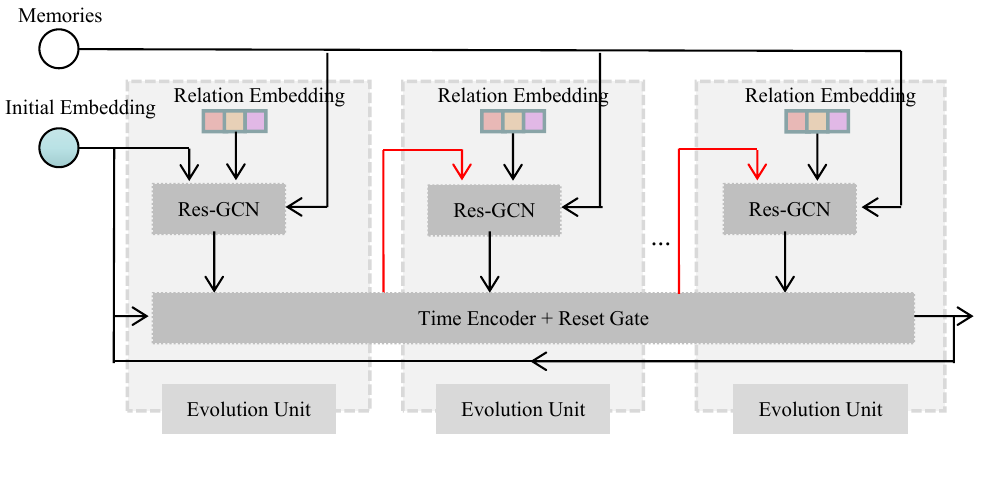}
	\end{center}
	\caption{This recurrent method uses the output of the time encoder and reset gating unit as the input of the next evolution unit, denoted as MTDM w. $\Box$.}
	\label{fig4}
\end{figure}

\subsubsection{\bfseries Structure-related variants}

For model structure-related variants, we pay attention to four aspects, w.o R$\_$Gate, w. $\bigstar$, w. $\Box$, and w. Tsf, where w. $\bigstar$ denotes replacing the Res-GCN in MTDM by the structural encoder in \cite{li2021temporal}, w. Tsf means replacing GRU in MTDM with the transformer encoder, w. $\Box$ represents using the proportion method in Fig. \ref{fig4}. Fig. \ref{fig4} uses the output of the time encoder and reset gating unit as the input of the next evolution unit. The formula of Fig. \ref{fig4} is,
\begin{equation}\label{eq3}
\vec{e}_{s, t}^{(w)} = \text{Res-GCN}(\mathbf{H}_{s,t-1}^{\text{TREN}}, \mathcal{F}_{t})
\end{equation}
Namely, replacing Eq. (\ref{eq2}) by Eq. (\ref{eq3}).

Ablation results for various structural variations are illustrated in Table \ref{tab4} and evaluated on two standard datasets, YAGO and ICEWS14.
We observe that R$\_$Gate slightly improves the prediction accuracy of the model on the ICEWS14 dataset.  

Compared with the default MTDM, MTDM w. $\bigstar$ causes a 2.6\% accuracy decrease on the YAGO dataset. However, only a 0.3\% performance decreases on the ICEWS14 dataset. This phenomenon proves the effectiveness of Res-GCN in predicting missing entities with short-term dependence, which also proves that the multi-hop aggregation will hurt early aggregation information.

Next, by replacing the GRU with the encoder of the transformer, we found that the prediction performance of the model decrease significantly.
Therefore, considering the memory consumption and method performance, MTDM applies the GRU as the time encoder.

By replacing the Eq. (\ref{eq2}) with the Eq. (\ref{eq3}) in Algorithm \ref{example}, we study the influence of different proportion modes on MTDM prediction performance.
It is worth noting that MTDM with the recurrent proportion mode in Fig. \ref{fig4} (w. $\Box$) is worse than the default MTDM. In our opinion, this is because insufficient memories introduce reasoning errors when it contributes to the temporal reasoning task. Meanwhile, the recurrent method will accumulate inference errors. That is, the dissolved facts are reactivated in the next evolution.

\begin{figure*}[t]
	\begin{center}
		
		\includegraphics[width=0.8\linewidth]{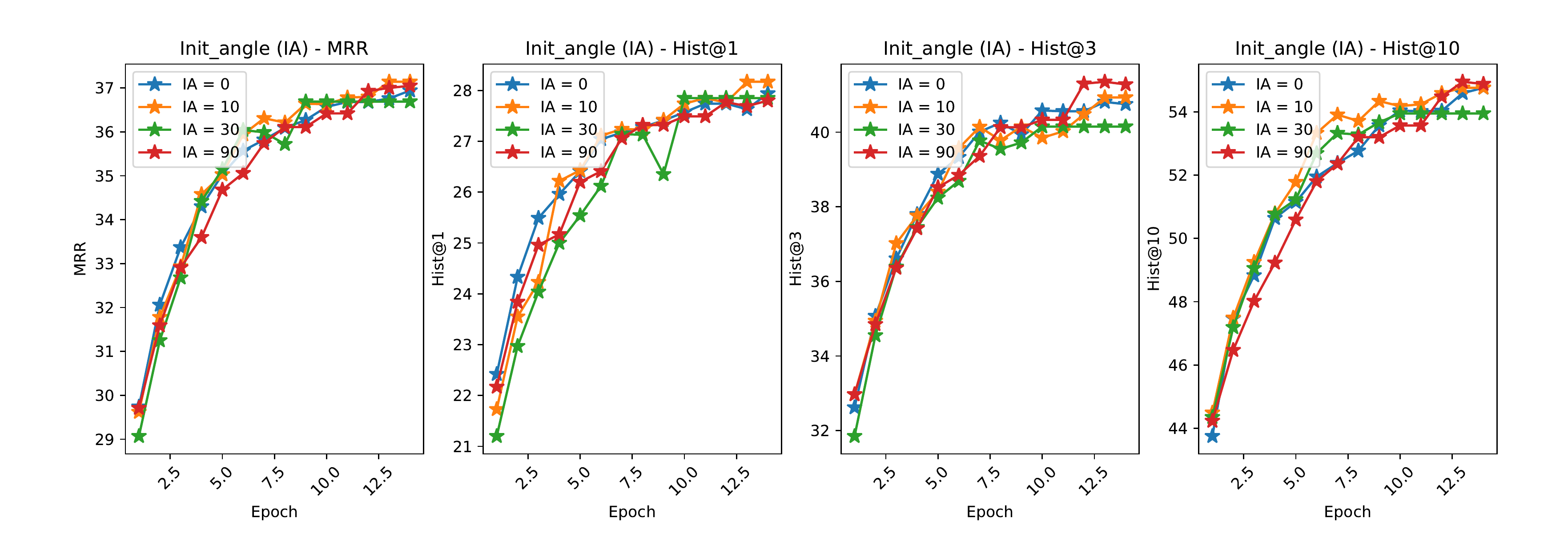}
	\end{center}
	\caption{The impact of the initial angle in Eq. (\ref{angle}) towards the reasoning performance of MTDM.}
	\label{fig5}
\end{figure*}

\begin{figure*}[t]
	\begin{center}
		
		\includegraphics[width=0.8\linewidth]{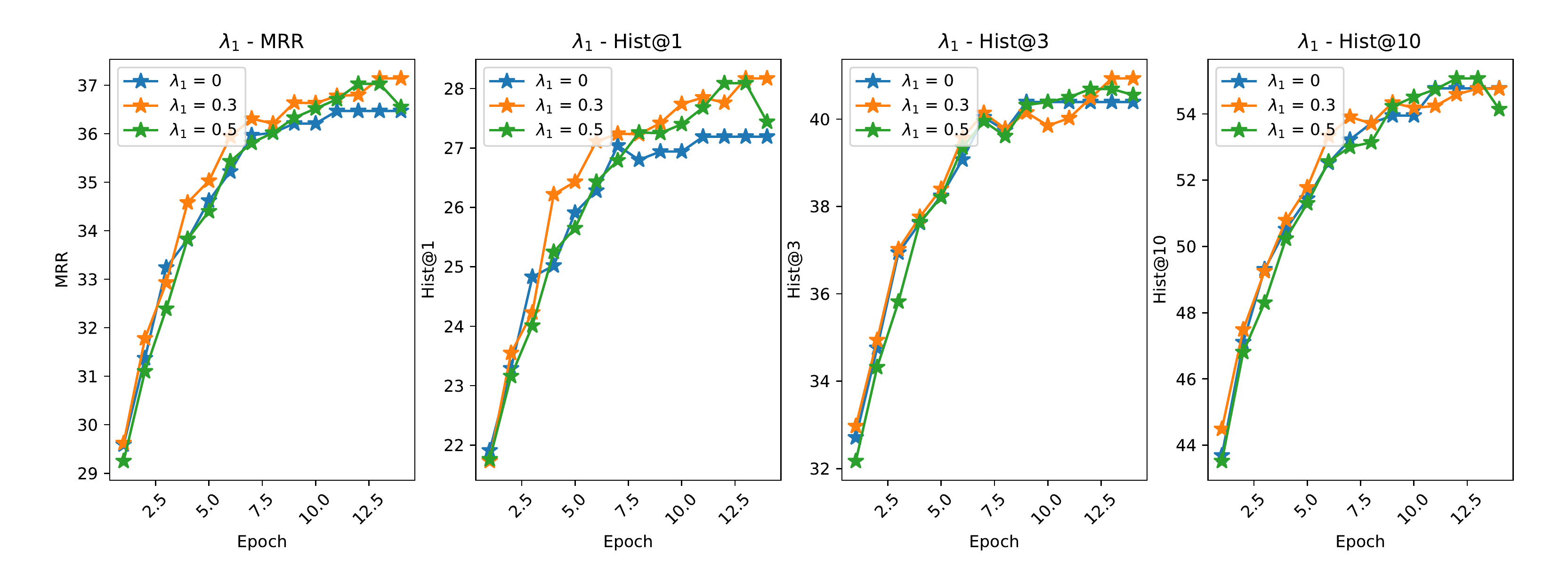}
	\end{center}
	\caption{The impact of the hyper-parameter $\lambda_{1}$ in Eq. (\ref{eq4}) towards the reasoning performance of MTDM.}
	\label{fig6}
\end{figure*}

\begin{figure*}[t]
	\begin{center}
		
		\includegraphics[width=0.8\linewidth]{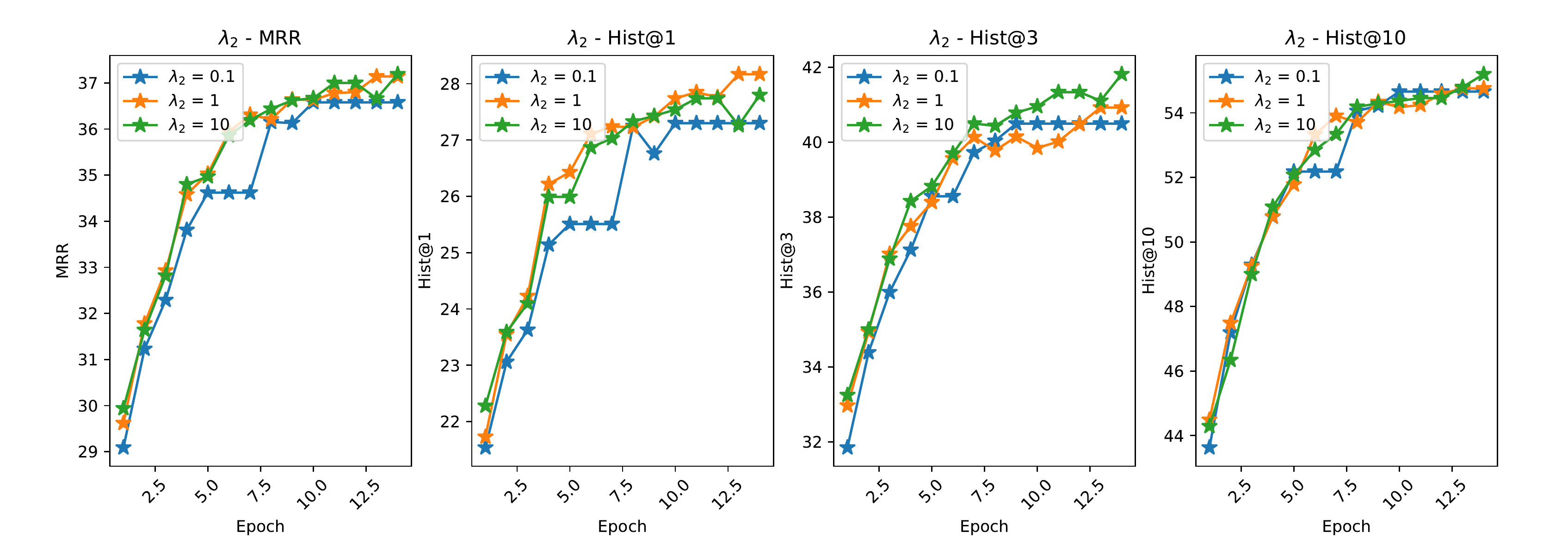}
	\end{center}
	\caption{The impact of the hyper-parameter $\lambda_{2}$ in Eq. (\ref{eq4}) towards the reasoning performance of MTDM.}
	\label{fig7}
\end{figure*}

\renewcommand{\arraystretch}{1.3} 
\begin{table*}[tp]

	\centering
	
	\fontsize{7}{8}\selectfont
	
	\begin{threeparttable}
		
		\caption{Case study. Several examples are used to illustrate the reasoning process.}
		
		\label{tab8}
		
		\begin{tabular}{c|ccc|c}
			
			\toprule
			
			Ground Truth Triplets &Mode & Predictions & Probability & Reason\cr
			
			\midrule
			
			\multirow{10}{*}{(South$\_$Korea,Intent intelligence,Japan?)}&\multirow{5}{*}{-} &North$\_$Korea &0.5328&(South$\_$Korea, Host a visit, North$\_$Korea)\cr
			&&China &0.0859&(South$\_$Korea, Consult, China)\cr
			&&Japan &0.0603&(South$\_$Korea, Express intent to meet or negotiate, Japan)\cr
			&&Mexico&0.0391&No direct links between South$\_$Korea and Mexico\cr
			&&Colombia &0.0364&No direct links between South$\_$Korea and Colombia\cr
			\cline{2-5}
			&\multirow{5}{*}{GT} &Japan &0.2955	&(South$\_$Korea, Intent intelligence, Japan)\cr
			&&China &0.1801&No direct links between South$\_$Korea and China\cr
			&&Thailand&0.1139&No direct links between South$\_$Korea and Thailand\cr
			&&	Pope$\_$Francis&0.0590&No direct links between South$\_$Korea and Pope$\_$Francis\cr
			&&	North$\_$Korea&0.0378&No direct links between South$\_$Korea and North$\_$Korea\cr
			\hline
			\multirow{10}{*}{(Japan,Intent cooperate,China?)}&\multirow{5}{*}{-} &South$\_$Korea &0.3685&(Japan, Engage in negotiation, South$\_$Korea)\cr
			&&China &	0.3473&(China, Intent cooperate, Japan)\cr
			&&North$\_$Korea &	0.0747&(North$\_$Korea, Criticize or denounce, Japan)\cr
			&&	France&	0.0627&No direct links between Japan and France\cr
			&&	Iran &	0.0170&No direct links between Japan and Iran\cr
			\cline{2-5}
			&\multirow{5}{*}{GT} &South$\_$Korea &0.5975&(Japan, Consult, South$\_$Korea)\cr
			&&China &	0.1356&(Japan, Consult, China)\cr
			&&  North$\_$Korea&	0.0785&No direct links between South$\_$Korea and North$\_$Korea\cr
			&&	Philippines&	0.0321&No direct links between South$\_$Korea and Philippines\cr
			&&	Vietnam&	0.0318&No direct links between South$\_$Korea and Vietnam\cr
			\hline
			\multirow{10}{*}{(North$\_$Korea, Criticize or denounce, Japan?)}&\multirow{5}{*}{-} &UN$\_$Security$\_$Council&0.3142&No direct links between North$\_$Korea and UN$\_$Security$\_$Council\cr
			&&  South$\_$Korea  &0.2883&No direct links between North$\_$Korea and South$\_$Korea\cr
			&&Japan  &	0.0832&No direct links between North$\_$Korea and Japan\cr
			&&	China	&	0.0223&No direct links between North$\_$Korea and China\cr
			&&	Head (South$\_$Korea) &0.0219&No direct links between North$\_$Korea and Head (South$\_$Korea)\cr
			\cline{2-5}
			&\multirow{5}{*}{GT} &South$\_$Korea &0.3242&(North$\_$Korea, Criticize or denounce, South$\_$Korea)\cr
			&& 	UN$\_$Security$\_$Council&0.1800&No direct links between South$\_$Korea and UN$\_$Security$\_$Council\cr
			&&  Barack$\_$Obama&0.1159&(North$\_$Korea, Criticize or denounce, Barack$\_$Obama)\cr
			&&	Japan&	0.0646&No direct links between South$\_$Korea and Japan\cr
			&&	Citizen$\_$(North$\_$Korea)&0.0393&No direct links between South$\_$Korea and Citizen$\_$(North$\_$Korea)\cr
			\hline
			\multirow{10}{*}{(China, Intent cooperate, Japan?)}&\multirow{5}{*}{-} &France&0.4799&(China, Intent cooperate, France)\cr
			&&  South$\_$Korea  &0.1704&No direct links between China and South$\_$Korea\cr
			&&	Japan&	0.0448&(China, Consult, Japan)\cr
			&&	Malaysia	&	0.0302&No direct links between China and Malaysia\cr
			&&	Philippines  &	0.0276&No direct links between Philippines and Philippines\cr
			\cline{2-5}
			&\multirow{5}{*}{GT} &South$\_$Korea &0.2719&No direct links between China and South$\_$Korea\cr
			&& 	Japan&0.2012&(China, Intent$\_$cooperate, Japan)\cr
			&&  Vietnam&	0.1550&No direct links between China and Vietnam\cr
			&&	Thailand&	0.0487&No direct links between China and Thailand\cr
			&&	Afghanistan&	0.0353&No direct links between China and Afghanistan\cr

			\bottomrule
			
		\end{tabular}
		
	\end{threeparttable}
	
\end{table*}

\subsubsection{\bfseries Data source-related variants}\label{dl}

To improve the understanding ability of MTDM towards events dissolution, We introduce dissolution learning constraint. We experimentally assign value for $\lambda_{4}$, $\lambda_{4}$ = 0.01, and denote it as MTDM w. ED. Experimental results are depicted in Table \ref{tab5}. We can observe that MTDM w. ED performs better than the default MTDM in the YAGO and ICEWS14 datasets but is worse than the default MTDM in the WIKI dataset. 
The insignificant improvement in TKGs inference indirectly proves that MTDM has the well cognitive ability to dissolved events. When removing the dissolution learning module, we find that the confidence of MTDM towards the training data is much greater than the dissolved data.
We believe that the poor performance of MTDM w. ED in WIKI is due to the strong correlation between several entities in the dataset. When some dissolved entities are selected as negative samples, they will affect the prediction accuracy of undissolved entities.

\subsubsection{\bfseries Various Hyper-parameters}

To be fair, we maintain values of other hyper-parameters and only modify the researched hyper-parameter. The initial angle (IA) in the similarity function Eq. (\ref{angle}) directly impacts the learned representations of TREN. That is, we constraint the minimum similarity between the evolutional representations $\mathbf{H}_{t}^{\text{TREN}}$ and the initial representations $\mathbf{H}_{<t_1}$. Comparative experiments are evaluated on ICEWS14 by setting IA to 0$^\circ$, 10$^\circ$, 30$^\circ$, 90$^\circ$. The results in Fig. \ref{fig5} indicate that the proposed MTDM realizes the best performance under IA = 10$^\circ$.

Fig. \ref{fig6} and Fig. \ref{fig7} illustrate the influence of hyper-parameters, $\lambda_{1}$, and $\lambda_{2}$, on the MTDM reasoning performance.
According to these experimental results, we set $\lambda_{1}$ and $\lambda_{2}$ to 0.3, 1.

\subsection{\bfseries Case Study}

We provide several case studies in Table \ref{tab8}. The first case (South$\_$Korea, Intent intelligence, Japan?) indicates that evaluation with $G_{<t_d}$ performs better than with $G_{<t_E}$. Historical facts in $G_{<t_E}$ tend to associate South$\_$Korea to North$\_$Korea, i.e. (South$\_$Korea, Host a visit, North$\_$Korea). However, there are directly related historical facts in $G_{<t_d}$, (South$\_$Korea, Intent intelligence, Japan).

Meanwhile, closely related entities affect the prediction results. For instance, both historical facts in $G_{<t_E}$ and $G_{<t_d}$ contain amounts of triplets (Japan, $*$, South$\_$Korea), as a result, the prediction probability of South$\_$Korea for (Japan, Intent cooperate, China?) is higher than of China even though there is directly related historical facts (Japan, Intent cooperate, China) in $G_{<t_E}$.

The third case indicates that recent facts will greatly affect the reasoning results even these intentions are not correct for the prediction, i.e. MTDM assigns South$\_$Korea and Barack$\_$Obama as the targets of (North$\_$Korea, Criticize or denounce, Japan?) for historical facts (North$\_$Korea, Criticize or denounce, South$\_$Korea) and (North$\_$Korea, Criticize or denounce, Barack$\_$Obama).

Finally, we observe that reasoning results are affected by the frequency of entities.
We list Top$\_$10 entities and their occurrence times summarized within recent historical facts, that is, (`Vietnam', 38), (`Afghanistan', 38), (`North$\_$Korea', 43), (`South$\_$Korea', 46), (`Thailand', 48), (`Japan', 67), (`Citizen$\_$(India)', 77), (`Iraq', 83), (`Iran', 136), (`China', 181).
Therefore, `Vietnam', `Afghanistan', `Thailand' are selected as the targets for the prediction of (China, Intent cooperate, Japan?) even there are no direct links between entities.

\begin{table}[tp]
	
	\centering
	
	\fontsize{6}{7}\selectfont
	
	\begin{threeparttable}
		
		\caption{The impact of memory related variations in MTDM, including without the time-aware recurrent evolutional network (w.o TREN), without deep memory (w.o DM), without transient learning network (w.o TLN). $\#$ means keeping other components of MTDM unchanged.}
		
		\label{tab3}
		
		\begin{tabular}{cc ccc ccc }
			
			\toprule
			
			\multicolumn{2}{c}{\multirow{2}{*}{MTDM}}&\multicolumn{3}{c}{YAGO}&
			
			\multicolumn{3}{c}{ICEWS14}\cr
			
			\cmidrule(lr){3-5}\cmidrule(lr){6-8}
			
			&Mode&MRR  &H@3&H@10&MRR&H@3&H@10\cr
			\midrule
			\multirow{5}{*}{\#} &- &{\bfseries 77.66}&{\bfseries 80.82}&{\bfseries 83.79 }&{\bfseries 37.12}&{\bfseries 41.00}&{\bfseries 54.52} \cr
			&w.o DM&77.66&80.82&83.79&35.38&39.23&53.11 \cr
			&w.o TREN&76.08&79.27&83.02&33.86&37.40&50.81 \cr

			&w.o TLN&74.04&76.62&80.64&37.12&41.00&54.52\cr


			\bottomrule
			
		\end{tabular}
		
	\end{threeparttable}
	
\end{table}

\renewcommand{\arraystretch}{1.3} 
\begin{table}[tp]

	\centering
	
	\fontsize{6}{7}\selectfont
	
	\begin{threeparttable}
		
		\caption{The impact of structural related variations in MTDM, including without reset gating units - w.o R$\_$Gate, different structural encoders - w. $\bigstar$, different time encoders - w. Tsf, and various recurrent modes - w. $\Box$.}
		
		\label{tab4}
		
		\begin{tabular}{cc ccc ccc }
			
			\toprule
			
			\multicolumn{2}{c}{\multirow{2}{*}{MTDM}}&\multicolumn{3}{c}{YAGO}&
			
			\multicolumn{3}{c}{ICEWS14}\cr
			
			\cmidrule(lr){3-5}\cmidrule(lr){6-8}
			
			&Mode&MRR  &H@3&H@10&MRR&H@3&H@10\cr
			
			\midrule

			\multirow{5}{*}{\#} &- &{ 77.66}&{ 80.82}&{ 83.79} &{\bfseries 37.12}&{ 41.00}&{ 54.52} \cr
			&w.o R$\_$Gate&{\bfseries 77.76}&{\bfseries 81.06}&{\bfseries 83.81}&36.73&40.79&{\bfseries 54.71} \cr
			&w. $\bigstar$&75.66&78.81&82.99&37.00&{\bfseries 41.02}&54.48 \cr
			
			&w. $\Box$&77.13&80.07&83.37&34.92&38.88&52.77\cr
			
			& w. Tsf &75.36&78.63&82.78     &35.61&39.52&53.55 \cr

			
			\bottomrule
			
		\end{tabular}
		
	\end{threeparttable}
	
\end{table}

\renewcommand{\arraystretch}{1.3} 
\begin{table*}[tp]

	\centering
	
	\fontsize{6}{7}\selectfont
	
	\begin{threeparttable}
		
		\caption{The impact of data source related variations in MTDM, which illustrates the effectiveness of events dissolution module.}
		
		\label{tab5}
		
		\begin{tabular}{cc cccc cccc cccc}
			
			\toprule
			
			\multicolumn{2}{c}{\multirow{2}{*}{MTDM}}&\multicolumn{4}{c}{YAGO}&\multicolumn{4}{c}{WIKI}&
			
			\multicolumn{4}{c}{ICEWS14}\cr
			
			\cmidrule(lr){3-6}\cmidrule(lr){7-10}\cmidrule(lr){11-14}
			
			&Mode&MRR &H@1 &H@3&H@10&MRR&H@1&H@3&H@10&MRR&H@1&H@3&H@10\cr
			
			\midrule
			
			\multirow{2}{*}{\#}&-  &{ 77.66}&73.85&{\bfseries 80.82}&{ 83.79} &{\bfseries63.18}&{\bfseries 60.24}&{\bfseries65.31}& {\bfseries67.89} &{ 37.12}&28.15&{ 41.00}&{\bfseries 54.52} \cr
			&w. ED&{\bfseries 77.79}&{\bfseries 74.00}&{ 80.61}&{\bfseries 83.86}&62.46 &59.32&64.71 &67.31 &{\bfseries 37.15}&{\bfseries 28.16}&{\bfseries 41.05}&{ 54.48} \cr

			
			\bottomrule
			
		\end{tabular}
		
	\end{threeparttable}
	
\end{table*}

\section{Conclusions}
In this paper, we propose a memory-triggered decision-making network (MTDM), which consists of a transient learning network and a time-aware recurrent evolutional network. 
Extensive experiments demonstrate the superior of MTDM on the extrapolation temporal reasoning task.
MTDM successfully alleviates its dependence on historical length and protects the multi-hop aggregation information by utilizing the residual structural encoder.
Additionally, the dissolution constraint helps MTDM understand events dissolution.
Since the transient learning network and the time-aware recurrent evolutional network jointly encodes entity attribute representations, MTDM spends less time than the best baselines on the training process.

\renewcommand{\arraystretch}{1.3} 

\bibliographystyle{IEEEtran}
\bibliography{ref}

\end{document}